%% file: iros_2025_jianwei-buzz.tex
\documentclass[letterpaper, 10 pt, conference]{ieeeconf}  

\IEEEoverridecommandlockouts                              

\usepackage[colorinlistoftodos]{todonotes}

\usepackage{caption}
\usepackage{subcaption}
\usepackage{amsmath}
\usepackage{amsfonts}
\usepackage{flushend}

\usepackage{graphicx}
\usepackage{booktabs}  
\usepackage{longtable} 
\usepackage{multirow} 
\usepackage{adjustbox}
\usepackage{comment}

\usepackage[hidelinks]{hyperref} 

\title{\LARGE \bf
Unreal Robotics Lab: A High-Fidelity Robotics Simulator with Advanced Physics and Rendering
}
\author{Jonathan Embley-Riches$^{*}$, Jianwei Liu$^{*}$, Simon Julier, and Dimitrios Kanoulas
\thanks{*equal contribution}
\thanks{The authors are with the Department of Computer Science, University College London, Gower Street, WC1E 6BT, London, UK. 
}
\thanks{This work was supported by the UKRI FLF [MR/V025333/1] (RoboHike).  For the purpose of Open Access, the author has applied a CC BY public copyright license to any Author Accepted Manuscript version arising from this submission.}}

\begin{document}
\makeatletter
\let\@oldmaketitle\@maketitle
\renewcommand{\@maketitle}{
    \@oldmaketitle
    \centering
    \vspace{-2.5pt}
        \hfill
        \includegraphics[width=0.325 \textwidth]{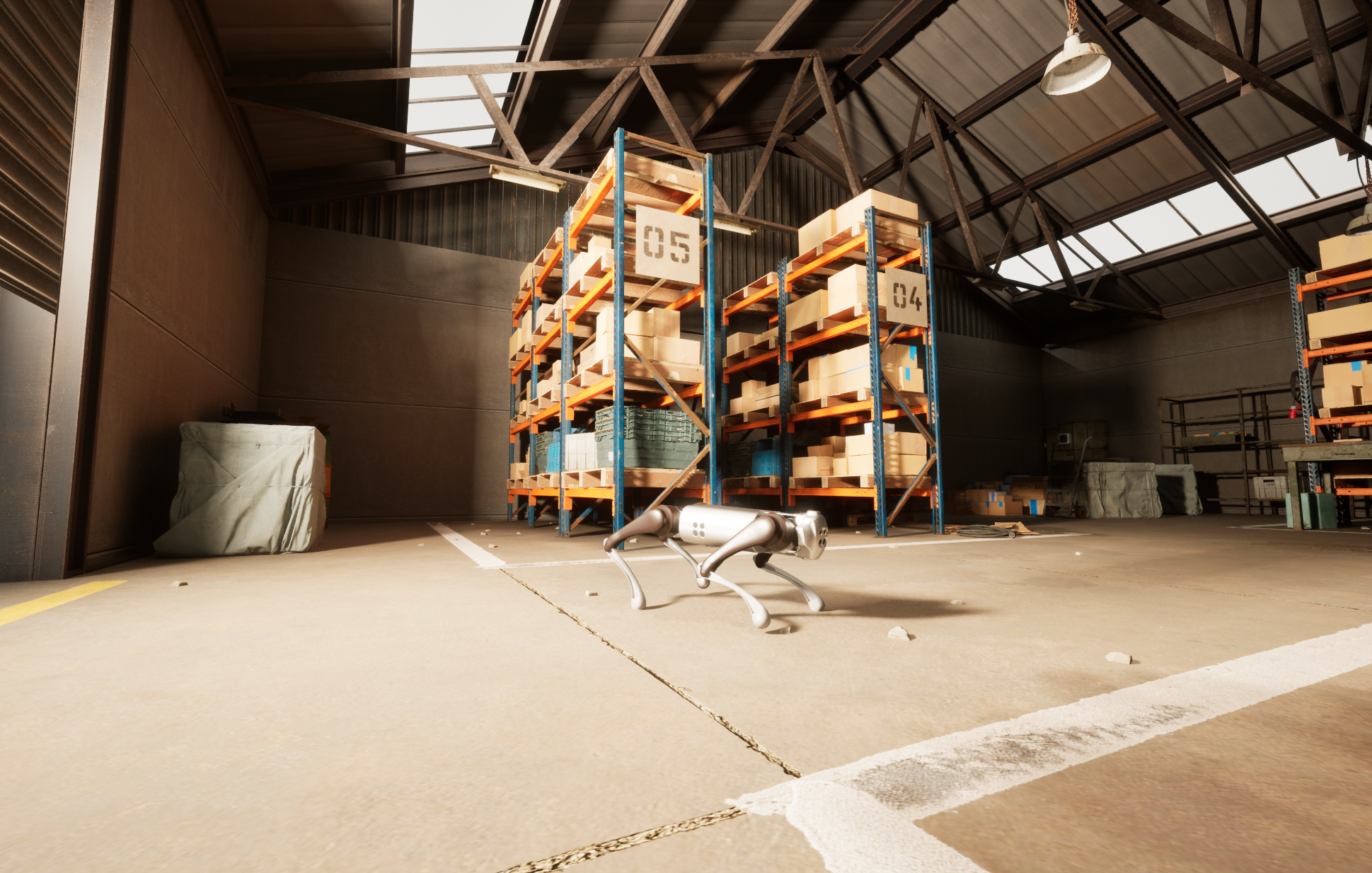}
        \includegraphics[width=0.325 \textwidth]{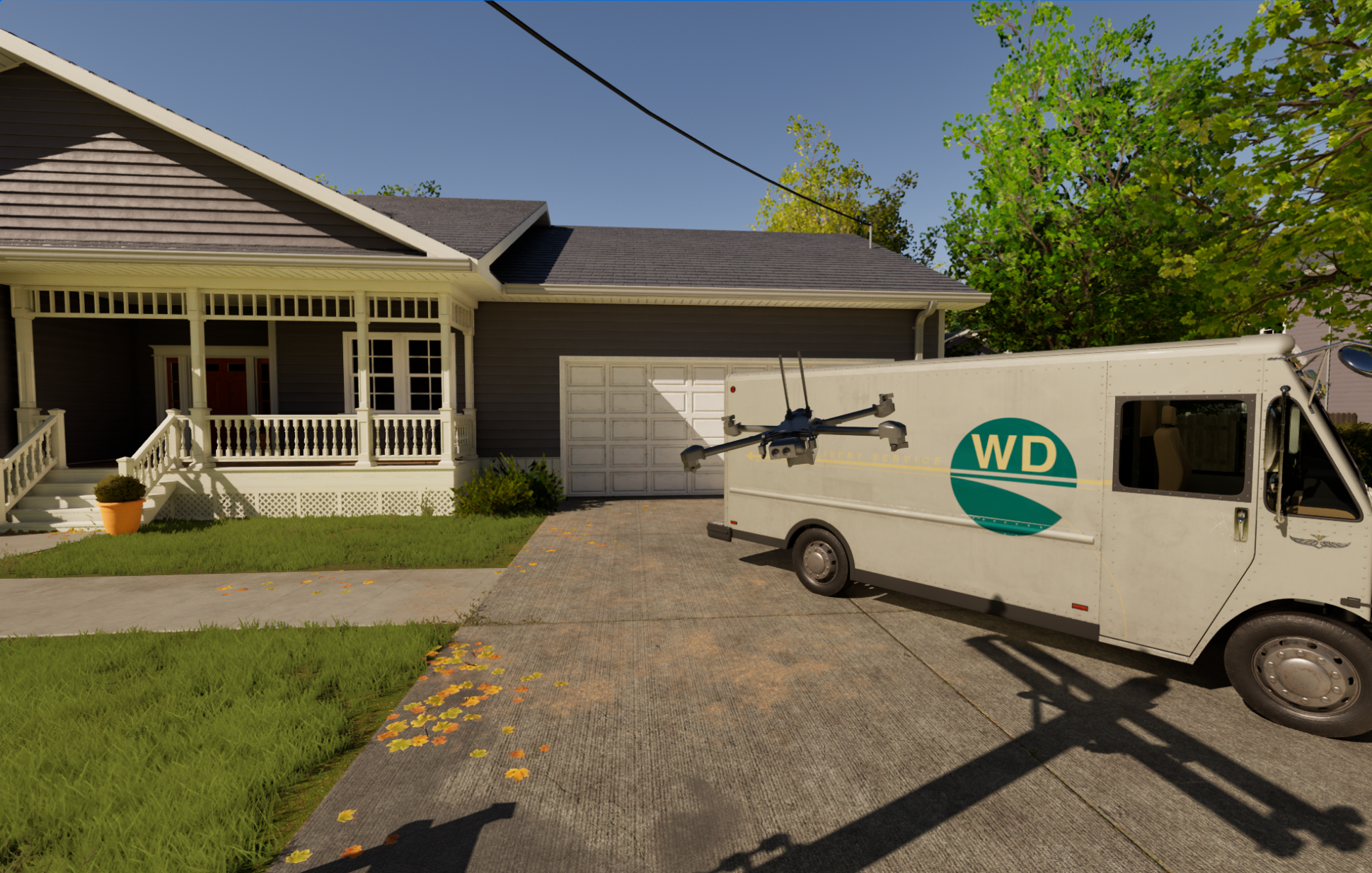}
        \includegraphics[width=0.325 \textwidth]{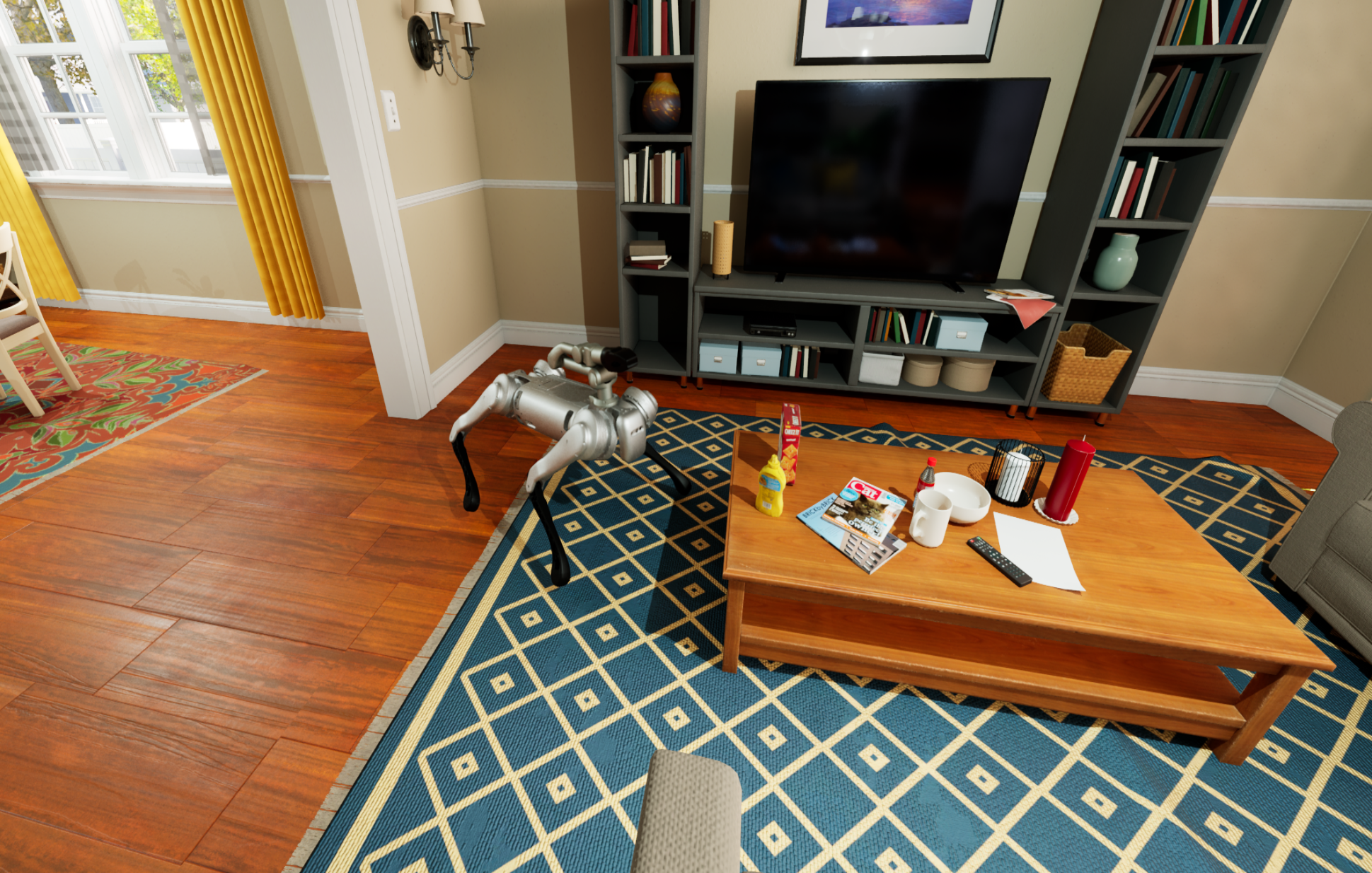}
        \hfill
        \vspace{0.1cm}
        \hfill
        \includegraphics[width=0.325\textwidth]{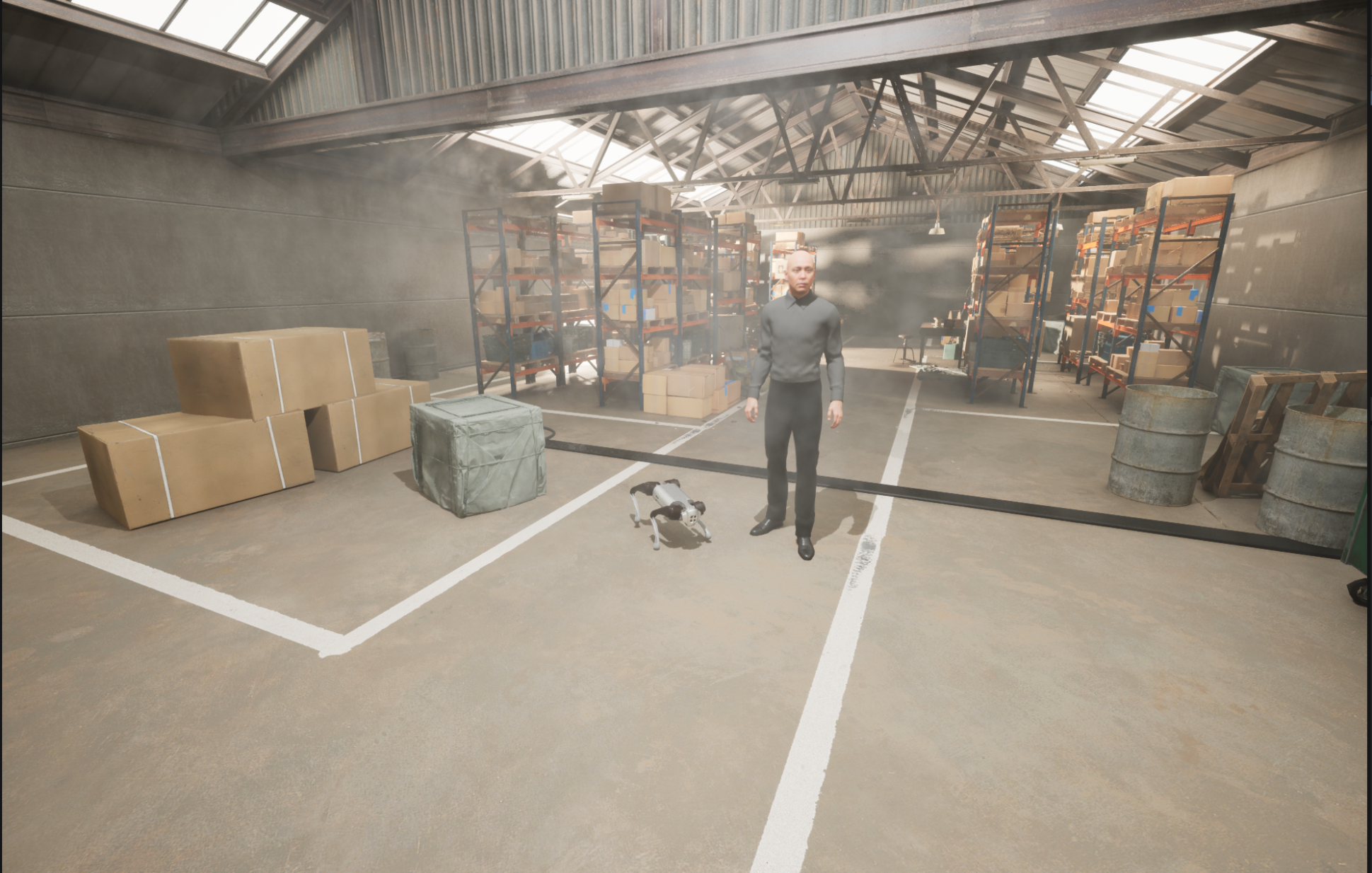}
        \includegraphics[width=0.325\textwidth]{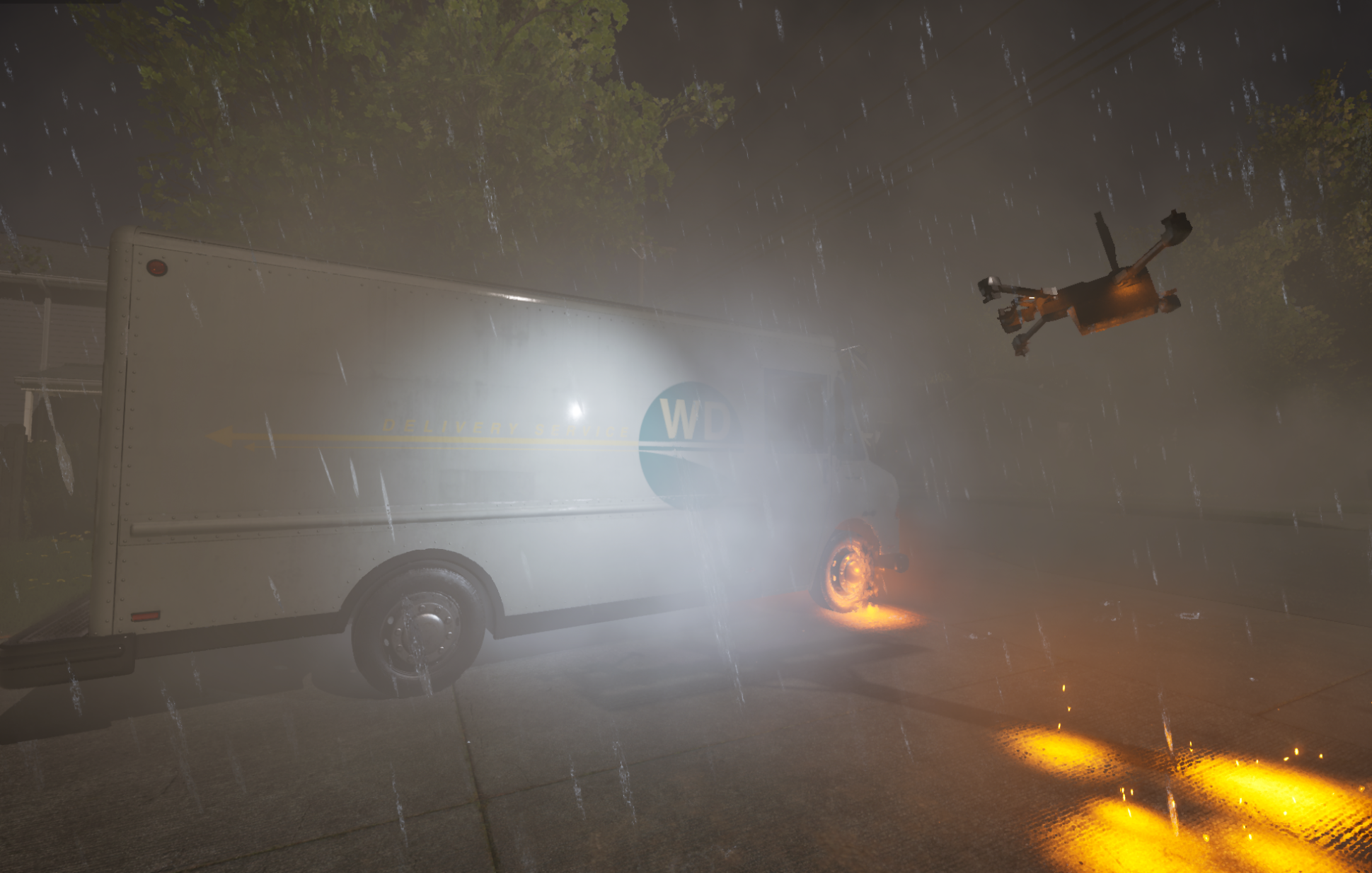}
        \includegraphics[width=0.325\textwidth]{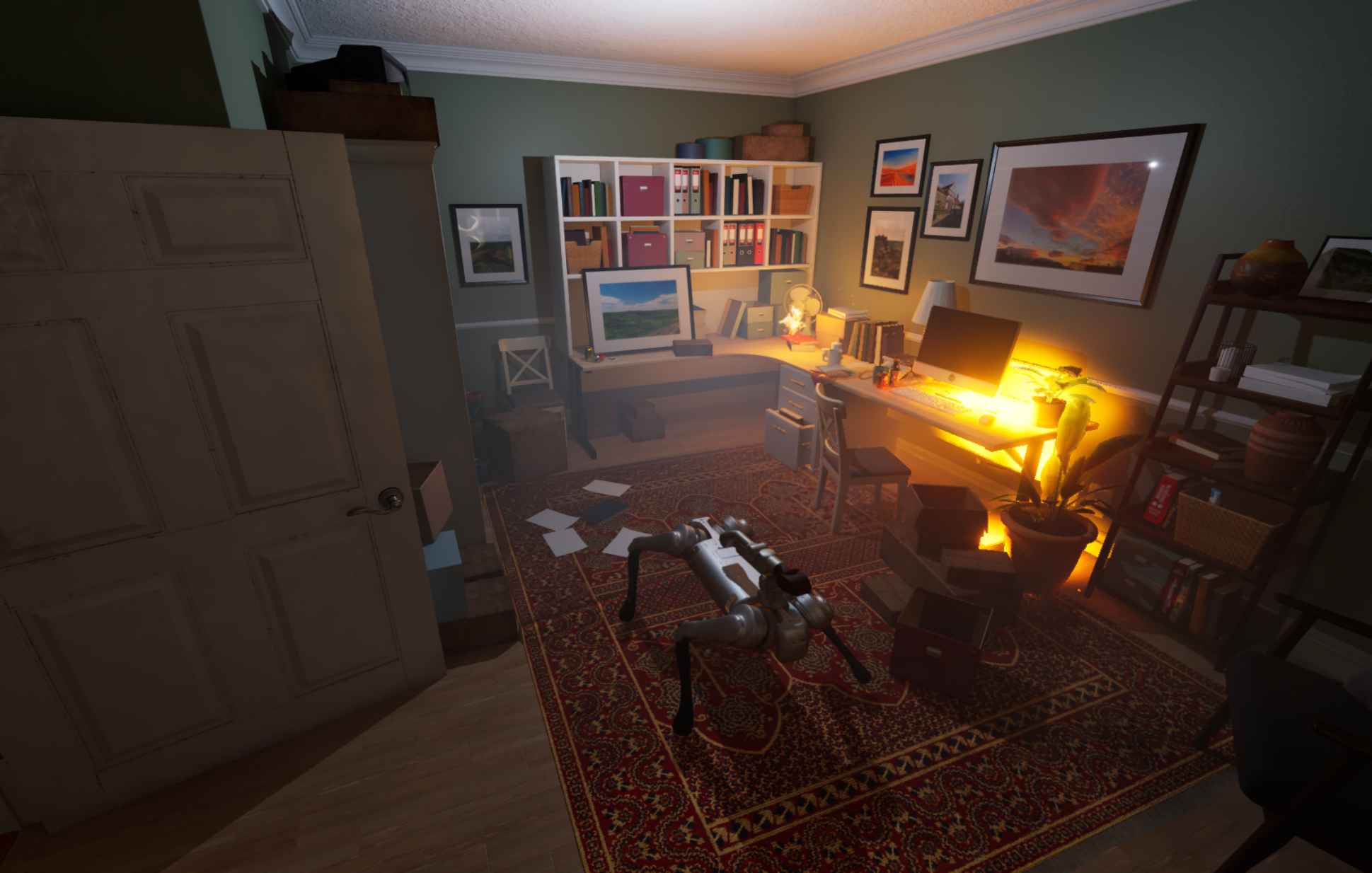}
        \hfill
    \captionof{figure}{Photorealistic renderings created by our simulation framework. Top row (types of robots): Unitree Go1 quadruped, Skydio X2 quadropter and Unitree B1-Z1 qudruped mobile manipulator Bottom row: various adverse visual conditions.}
    \label{fig:example_photo_realistic_render}%
    \vspace{-10.5pt}
}


\makeatother
\maketitle
\thispagestyle{empty}
\pagestyle{empty}


\begin{abstract}
High-fidelity simulation is essential for robotics research, enabling safe and efficient testing of perception, control, and navigation algorithms. However, achieving both photorealistic rendering and accurate physics modeling remains a challenge. This paper presents a novel simulation framework, the Unreal Robotics Lab (URL), that integrates the advanced rendering capabilities of the Unreal Engine with MuJoCo’s high-precision physics simulation. Our approach enables realistic robotic perception while maintaining accurate physical interactions, facilitating benchmarking and dataset generation for vision-based robotics applications. The system supports complex environmental effects, such as smoke, fire, and water dynamics, which are critical to evaluating robotic performance under adverse conditions. We benchmark visual navigation and SLAM methods within our framework, demonstrating its utility for testing real-world robustness in controlled yet diverse scenarios. By bridging the gap between physics accuracy and photorealistic rendering, our framework provides a powerful tool for advancing robotics research and sim-to-real transfer. Our open-source framework is available at~\url{https://unrealroboticslab.github.io/}.
\end{abstract}

\input{01_intro.tex}
\input{02_rw.tex}
\input{new_method_3}
\input{04_experiments}
\input{05_conclusion}


\addtolength{\textheight}{0cm}   

\bibliographystyle{IEEEtran}
\bibliography{IEEEabrv, iros_2025_jianwei-buzz}

\end{document}

%% file: 01_intro.tex
\section{INTRODUCTION}
Simulation is essential for robotics research, enabling safe, efficient development, testing, and validation of robotic systems before real-world deployment. High-fidelity environments must simultaneously deliver realistic perception and accurate physics interactions—an open challenge due to competing computational demands and the complexity of real-world dynamics~\cite{szot2021habitat}.

We introduce the \textbf{Unreal Robotics Lab (URL)}, a novel open-source framework that tightly integrates Unreal Engine’s state-of-the-art rendering pipeline (Lumen global illumination, Nanite virtualized geometry, Niagara particles)~\cite{unrealengine} with MuJoCo’s validated high-precision physics engine~\cite{todorov2012mujoco}. By embedding MuJoCo natively inside the Unreal ecosystem via shared-memory execution, URL achieves deterministic physics while exposing Unreal’s full authoring, environmental-effect, and tools directly to roboticists.

A key capability is the effortless generation of complex dynamic phenomena --- smoke, fire, water waves, dynamic lighting, and moving agents—that are rare or hazardous to capture in real datasets. These effects enable controlled benchmarking of vision-based policies under out-of-distribution conditions and the creation of large-scale training data for perception and navigation.


Unlike domain-specific simulators (e.g., autonomous driving or aerial robotics) or platforms focused on limited embodied-AI scenarios, URL is designed for \textbf{general-purpose robotic applications}, supporting quadrupeds, mobile manipulators, UAVs, and arms. While simulators such as Isaac Sim~\cite{NVIDIA_Isaac_Sim} offer broad capabilities, URL prioritises physical fidelity by retaining MuJoCo’s superior accuracy and stability in contact-rich tasks and incorporates \textbf{CoACD-based convex decomposition}~\cite{wei2022coacd} for improved non-convex object representation.

Our key contributions are:

\begin{itemize}
    \item \textbf{Native ``Unreal-First'' Architecture:} We introduce a shared-memory execution model that embeds the MuJoCo engine directly within the Unreal Engine environment. This eliminates the latency and synchronization overhead of traditional client-server models.

    \item \textbf{High-Fidelity Physics-to-Visual Synchronization:} By mirroring the MuJoCo schema within the UE Actor-Component hierarchy, we enable seamless synchronization between MuJoCo’s stable physics and UE’s photorealistic rendering pipeline (Lumen, Nanite, Niagara).

    \item \textbf{Automated Environmental Processing and Geometry Conversion:} The framework provides a robust pipeline for bridging complex UE environmental assets to physics-ready formats, utilizing multi-threaded terrain sampling and automated CoACD-based conversion for arbitrary static meshes.

    \item \textbf{Comprehensive ROS-Compatible Middleware:} A unified interface for sensors and actuators that supports diverse robotic platforms with native support for ROS and ZeroMQ communication. This enables the ``plug-and-play'' deployment of modern planners (e.g., ViNT, NoMAD) and SLAM frameworks (e.g., ORB-SLAM3) directly within the simulation.

    \item \textbf{Generalized Benchmarking for Diverse Conditions:} We provide a structured framework for evaluating robotic systems under rare or extreme environmental conditions, complemented by a deterministic replay system that allows for robust domain randomization testing across varied visual environments.
    
\end{itemize}

%% file: 02_rw.tex
\vspace{-1.5pt}
\section{RELATED WORK}\label{sec:rw}

Robotic simulation has advanced significantly in recent years, addressing key challenges in physics accuracy, photorealistic rendering, and real-time performance. This section reviews related work in three main areas: robotics-focused simulation platforms, physics engine comparisons, and benchmarking of SLAM and navigation under adverse visual conditions.

Game engines and simulation platforms offer photorealistic rendering~\cite{mittal2023orbit, puig2023habitat, dosovitskiy2017carla}, yet Unreal Engine provides distinct advantages through Lumen global illumination and Nanite virtualized geometry~\cite{unrealengine}. Lumen enables real-time global illumination and reflections without precomputed lighting—ideal for dynamic environments—while Nanite efficiently renders highly detailed assets with millions of triangles. Combined with its extensible ecosystem (third-party plugins, visual scripting, AI-driven NPCs, and cross-platform deployment), Unreal Engine is particularly well-suited for vision-based robotics research.

\subsection{Robotics-Focused Simulation Platforms}

Several simulators target robotic perception, control, and reinforcement learning~\cite{ellis2022navigation, shanks2025dreamernav}, each with different trade-offs in realism and efficiency. Gazebo~\cite{koenig2004design} is a widely-used open-source platform that supports multiple physics engines (ODE~\cite{koenig2004design}, Simbody~\cite{sherman2011simbody}, Bullet~\cite{coumans2016pybullet}, DART~\cite{lee2018dart}), although its rendering lags behind modern game-engine solutions. MuJoCo~\cite{todorov2012mujoco} excels in accurate, efficient physics for control tasks but lacks built-in photorealistic rendering. GPU-accelerated frameworks such as Isaac Orbit~\cite{mittal2023orbit} and Isaac Gym~\cite{makoviychuk2021isaac} prioritise speed for large-scale reinforcement learning, while Isaac Sim~\cite{NVIDIA_Isaac_Sim} (built on NVIDIA Omniverse) uses PhysX~\cite{nvidia_physx} for scalability at the expense of contact-dynamics accuracy compared with MuJoCo. Intel SPEAR~\cite{simulator_intelspear} provides photorealistic indoor environments for embodied AI, and domain-specific simulators such as CARLA~\cite{dosovitskiy2017carla} and AirSim~\cite{shah2018airsim} leverage Unreal Engine for urban/aerial scenarios. Manipulation-focused suites RoboSuite~\cite{robosuite2020} and RoboCasa~\cite{robocasa2024} rely on MuJoCo physics but offer limited environmental complexity.

Although most platforms emphasise either physics fidelity or visual realism, few achieve both. URL bridges this gap by natively integrating MuJoCo’s precise physics with Unreal Engine’s state-of-the-art rendering in an editor-first, ROS-native framework designed for general-purpose robotics (quadrupeds, mobile manipulators, UAVs, and arms).

\subsection{Physics Engines for Robotics Simulation}

Physics engines are critical for realistic contact, soft-body, and fluid modelling. MuJoCo~\cite{todorov2012mujoco} is widely favored for its for soft-contact accuracy and stability in complex robotic tasks~\cite{erez2015simulation}. ODE (used in Gazebo) suffers from stability issues in high-DOF scenarios and requires extensive tuning. Bullet~\cite{coumans2016pybullet} is popular for reinforcement learning and manipulation but lags MuJoCo in precision. NVIDIA PhysX~\cite{nvidia_physx} offers GPU acceleration suitable for fast RL yet omits Coriolis forces and exhibits lower linear-stability accuracy. Unreal Engine’s Chaos and Havok solvers~\cite{havok_physics} excel at real-time destruction and particle effects but are not optimised for high-precision joint dynamics or soft-contact robotics.

The continued adoption of MuJoCo’s physics is further evidenced by the fact that next-generation GPU-accelerated simulators such as \textbf{Genesis}~\cite{Genesis} and \textbf{Newton}~\cite{newton_physics} have adopted \textbf{MJWarp}~\cite{mujoco_warp} (a high-performance GPU backend directly derived from MuJoCo) as their core physics engine. This widespread use in the latest state-of-the-art platforms confirms that MuJoCo’s contact modelling, accuracy, and stability remain unmatched—even as the field shifts toward massive parallel GPU simulation.


URL therefore retains native MuJoCo as the sole physics backend while inheriting Unreal Engine’s advanced rendering, particle systems, and tools—delivering both photorealism and validated physical fidelity without compromise.

\subsection{Benchmarking SLAM and Navigation Under Adverse Visual Conditions}

Evaluating SLAM and navigation algorithms under adverse conditions (smoke, fog, dynamic lighting, or moving agents) remains difficult, as most public datasets and simulators capture only structured, well-lit scenes~\cite{EmbleyRiches2025SAGESA}. Envodat~\cite{nwankwo2024envodat} provides one challenging real-world dataset, yet scale and diversity are still limited. URL addresses this limitation by synthetically generating controllable rare-event scenarios via Niagara particles and fluids, together with dynamic NPCs and vehicles scripted through Unreal’s NavMesh and Behavior Trees.

Current simulators typically prioritise either high-precision physics (e.g., MuJoCo-based) or photorealistic rendering (e.g., CARLA, AirSim). While our work shares the high-level goal of SPEAR~\cite{simulator_intelspear} in combining Unreal rendering with MuJoCo physics, the design philosophies differ fundamentally. SPEAR is structured as a Python-based OpenAI Gym environment for task-specific embodied-AI research. In contrast, URL is built for \textit{systems-level} validation: it offers native ROS integration, direct scene authoring inside the Unreal Editor (with automatic asset processing and CoACD convex decomposition), and seamless import of any MuJoCo-compatible robot. This enables full-stack testing of complete robotic software pipelines exactly as they would be deployed in the real world, together with repeatable benchmarking of perception and planning robustness in behaviorally rich, non-static environments.

%% file: new_method_3.tex
\section{Simulator Architecture}
\label{sec:method}

\subsection{System Overview}
\label{subsec:method_overview}
\begin{figure}[t]
    \centering
    \includegraphics[width=\columnwidth]{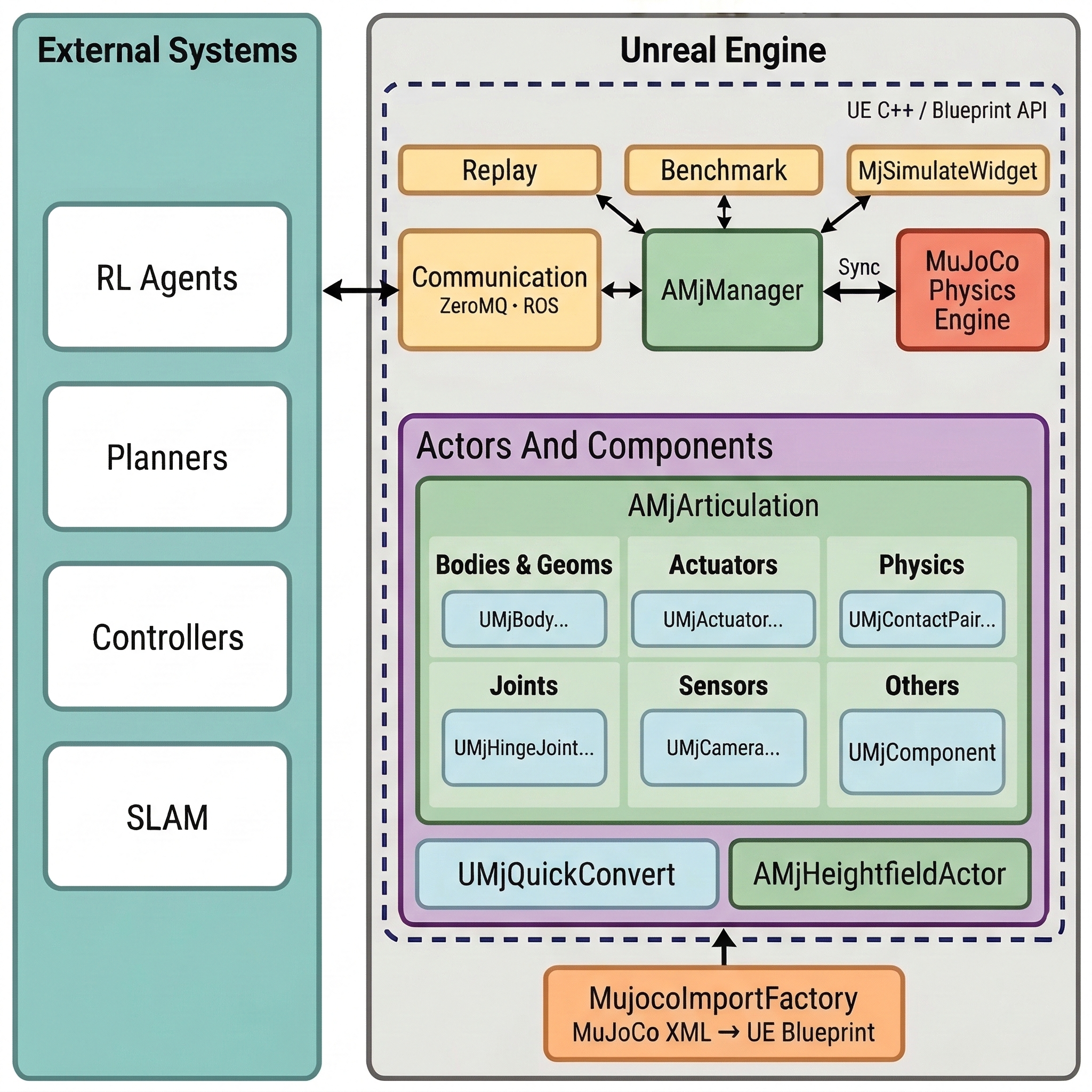}%
    \caption{Unreal - MuJoCo Simulator system diagram.}
    \label{fig:simulation_sys_diagram}%
\end{figure}

We propose an ``Unreal Engine-first'' simulation framework (Fig.~\ref{fig:simulation_sys_diagram}) that embeds the MuJoCo physics engine entirely within the Unreal Engine (UE) ecosystem. By discarding traditional client-server communication models in favor of in-process, shared-memory execution, the architecture ensures absolute physical determinism. This design enables the physics simulation to natively leverage UE's photorealistic rendering pipeline (Lumen, Nanite), environmental effects (Niagara), and AI navigation (NavMesh) without the latency or synchronization overhead inherent in distributed architectures.

\subsection{World Authoring and Asset Instantiation}
\label{subsec:method_authoring}

A core principle of this framework is the comprehensive mirroring of the MuJoCo schema within UE's Actor-Component hierarchy. Rather than providing an abstracted wrapper, the system exposes nearly the entire MuJoCo specification---including global defaults, solver options, and runtime configurations---as native, strongly-typed properties directly within the UE interface. The framework caters to both legacy MuJoCo users and native Unreal Engine development through a tiered ingestion pipeline.

Users can integrate legacy models by dragging standard MuJoCo XML files directly into the UE Content Browser. This triggers a dedicated factory that parses the MJCF specification to automatically generate a structurally equivalent UE Blueprint. Articulated robots and environmental objects are instantiated as \texttt{AMjArticulation} Actors, which manage a recursive tree of localized components such as \texttt{UMjBody}, \texttt{UMjGeom}, \texttt{UMjJoint}, and \texttt{UMjTendon}. This process automatically imports and converts all referenced meshes and primitives into native UE assets, creating a persistent representation that benefits from UE's advanced material and rendering systems.

For native authoring, users can instantiate an empty \texttt{AMjArticulation} and procedurally build the hierarchy through the standard UE ``Add'' interface. The architecture abstracts away from ID-based indexing and manual pointer arithmetic; users can access live physics data directly through the articulation's component references without managing raw \texttt{mjModel} or \texttt{mjData} addresses. Scene construction is finalized by placing these Blueprints within a level containing an \texttt{AMjManager} Actor. Upon initialization, the manager performs a scene-wide scan to aggregate all active Actors into a contiguous specification. To optimize performance, the framework utilizes asynchronous asset serialization and caching for mesh generation, ensuring complex geometries are processed once. The system maintains bidirectional compatibility, allowing users to export UE-authored levels back to standardized MuJoCo formats.

\subsection{Environmental Sampling and Geometry Processing}
\label{subsec:method_env}
\begin{figure}[ht!]
    \centering
    \subfloat[Unreal Engine Landscape]{%
        \includegraphics[width=0.49\columnwidth]{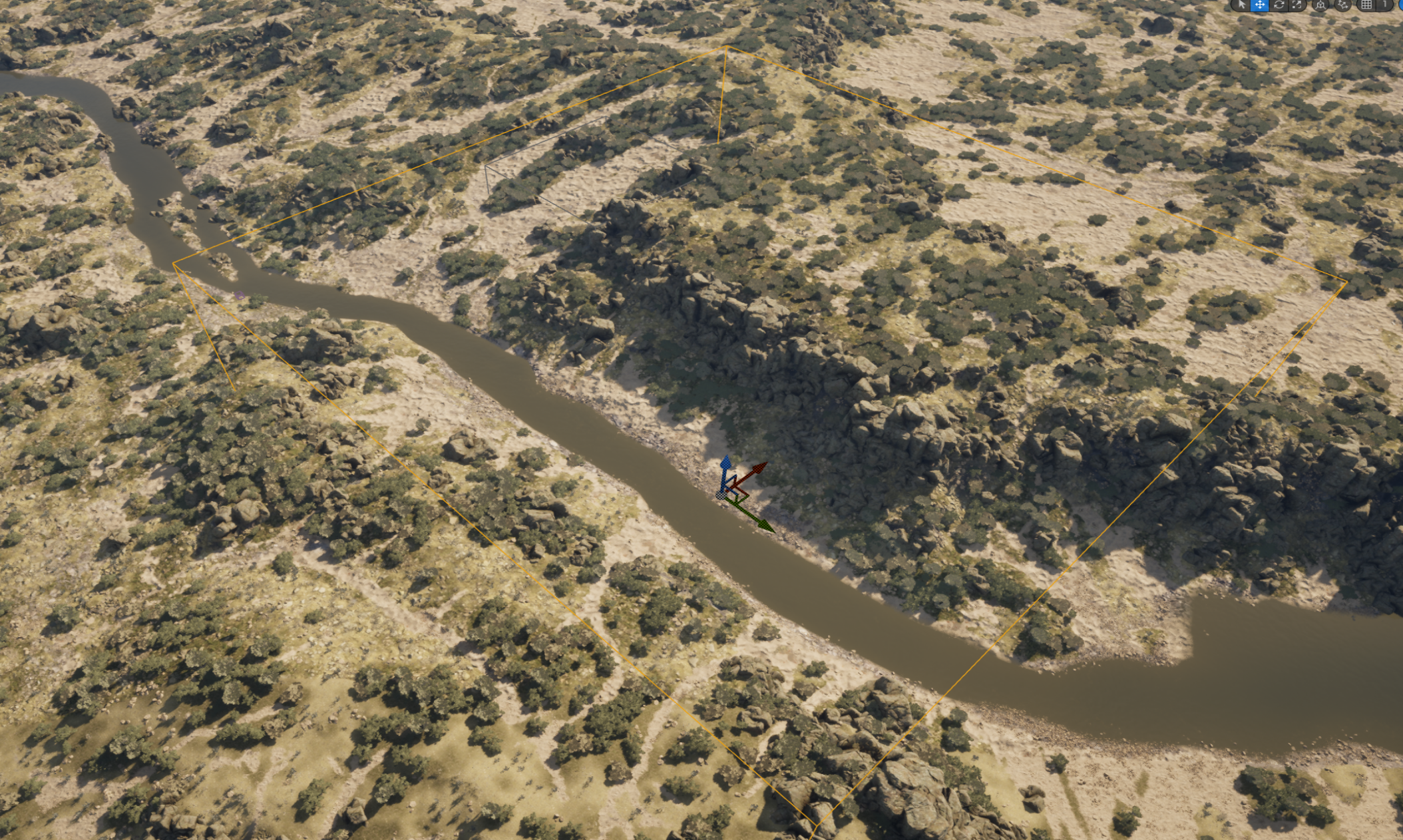}
        \label{subfig:UELandscape}
    } 
    \subfloat[MuJoCo HeightField]{%
        \includegraphics[width=0.49\columnwidth]{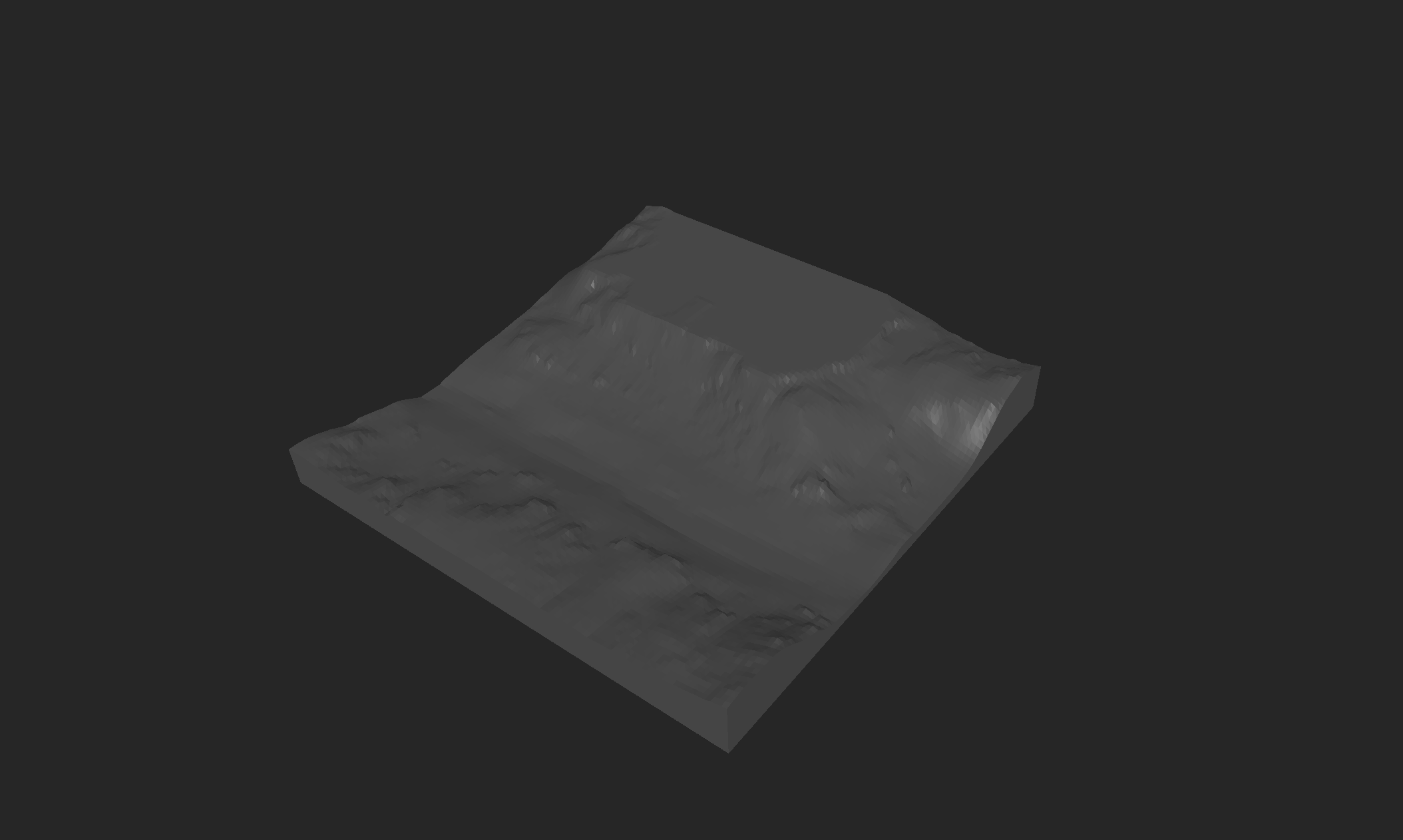}
        \label{subfig:MJHeightFIeld}
    }
    \caption{Unreal Engine Landscape conversion to MuJoCo HeightField.}
    \label{fig:LandscapeConversion}%
    \vspace{-15.5pt}
\end{figure}

To bridge the gap between UE’s diverse environmental assets and MuJoCo’s collision requirements, the framework provides automated conversion utilities. The \texttt{AMjHeightfieldActor} facilitates terrain interaction by executing multi-threaded, bounded raycasts to sample the environment. This actor is not restricted to UE Landscapes; it can sample any geometry not already managed by the MuJoCo solver, specifically any Actor lacking an \texttt{AMjArticulation} or a \texttt{UMjQuickConvertComponent}. This allows for the generation of optimized MuJoCo heightfields from complex, non-articulated environmental geometry while minimizing memory overhead (Fig.~\ref{fig:LandscapeConversion}). For discrete objects, the \texttt{UMjQuickConvertComponent} enables the instant conversion of arbitrary UE static meshes into MuJoCo-compatible collision bodies using CoACD~\cite{wei2022coacd} for automatic convex hull decomposition.

\subsection{Unified Interaction and Bi-Directional Physics}
\label{subsec:method_sync}
To maintain strict determinism, the MuJoCo simulation executes on a dedicated asynchronous thread. The solver steps at the specific timestep defined within the MuJoCo model, which is exposed and configurable via the \texttt{AMjManager}. Interaction is unified across the framework's API; every MuJoCo property, function, and sensor is accessible via both C++ and UE Blueprints. This includes a comprehensive sensor interface where all MuJoCo-native sensors are exposed as UE components, providing researchers with Blueprint nodes to access real-time data streams. Furthermore, the framework includes a native graphical UI (\texttt{MjSimulateWidget}) that replicates the functionality of MuJoCo's standalone \textit{Simulate}, allowing for real-time visualization, state monitoring, and interactive control within the UE viewport. While the framework handles pointer management, users retain the ability to interact with the raw MuJoCo structs if required for custom solver logic. This flexibility is complemented by a robust one-way kinematic coupling (\textit{mocap}) which allows UE's native Chaos physics and AI systems to influence the MuJoCo solver in real-time. For example, an NPC governed by the UE NavMesh can be assigned a \texttt{UMjQuickConvertComponent} with its ``Unreal Driven'' property enabled, transforming it into a \textit{mocap} body that moves in synchronization with UE's physics.

\subsection{Communication, Metrics, and Replay}
\label{subsec:comms_bench}
The framework supports multiple communication modalities, allowing users to choose between a modified ROS plugin~\cite{mania19scenarios} or a high-performance ZeroMQ (ZMQ) bridge. These plugins automate the mapping of MuJoCo and UE sensors to their respective topics or sockets based on the robot's configuration. For empirical evaluation, a \texttt{MetricManager} provides a structured framework for recording time-series data directly to CSV. Furthermore, a high-fidelity Replay System caches joint kinematics and sensor streams, enabling researchers to record physical trajectories and deterministically replay them across iteratively augmented visual environments for robust domain randomization testing.

\subsection{Example Deployment of Robots, Controllers, and Visual Planners}\label{subsec:method_deployment}
The system provides a unified framework in which robots, controllers, and planners operate as in real-world deployments. All components communicate through ROS, ensuring seamless integration with external systems.
\subsubsection{Example Robot Integrations} The framework supports quadrupeds, quadruped mobile manipulators, UAVs, and robotic manipulators, enabling diverse experimental setups for locomotion, aerial navigation, and manipulation.

\subsubsection{Controllers}\label{subsec:method_controllers} The system includes joint controllers based on position and torque, as well as more advanced controllers such as Walk-These-Ways~\cite{margolis2023walk} and VBC~\cite{liu2024visual}. UAVs use a PID-based controller, whereas manipulators rely on inverse kinematics (IK) methods such as Mink~\cite{Zakka_Mink_Python_inverse_2024}. Since controllers operate within the ROS ecosystem, new control strategies can be easily integrated.

\subsubsection{Planners}\label{subsec:method_planners} Motion planning is handled using learning-based and optimization-based planners such as ViNT~\cite{shah2023vint}, GNM~\cite{shah2023gnm}, and NoMAD~\cite{sridhar2024nomad}. The ROS-based architecture of the framework ensures compatibility with additional external planners.

\subsubsection{Visual SLAM} The system includes visual SLAM techniques such as OrbSLAM2~\cite{murORB2}, OrbSLAM3~\cite{campos2021orb}, and MASt3R-SLAM~\cite{murai2024mast3r}, enabling real-time localization and mapping. These methods integrate directly into the framework, enabling realistic perception and mapping in simulated environments.

%% file: 04_experiments.tex
\section{Experiments}\label{sec:exp}

\subsection{Simulation and Real-World Comparision}
To evaluate the alignment between simulated and real-world visual inputs for image encoders, we employ Grad-CAM~\cite{selvaraju2017grad}, EigenCAM~\cite{muhammad2020eigen}, cosine similarity and KL Divergence~\cite{kullback1951information}. These methods assess both spatial attention and feature distribution similarity. Specifically, spatial attention is computed using Grad-CAM and EigenCAM in equations~\ref{eq:eq1} and~\ref{eq:eq2} respectively to assess feature distribution similarity. \textbf{Grad-CAM}  (Gradient-weighted Class Activation Mapping) computes gradients of class score $  y^c  $ (pre-softmax logit for class $  c  $) w.r.t. feature maps $  A^k  $ (spatial activations from channel $  k  $ in the final conv layer):
\begin{equation}
    \alpha_k^c = \frac{1}{Z} \sum_{i} \sum_{j} \frac{\partial y^c}{\partial A^k_{ij}}, \quad
    L^c = \text{ReLU} \left( \sum_k \alpha_k^c A^k \right)
    \label{eq:eq1}
\end{equation}
where  $ \alpha_k^c  $ is the Global Average Pooling (GAP)-weighted importance of channel $k$ for class $c$ ($i,j$: spatial indices; $Z$: width $\times$ height). Heatmap $L^c$ is the weighted sum of maps with ReLU, yielding a saliency map of regions influencing the model's class $c$ prediction.

\textbf{EigenCAM} generates class-agnostic saliency maps via principal component analysis (PCA):
\begin{equation}
    C = \frac{1}{Z} \sum_{i,j} (A^k_{ij} - \mu) (A^k_{ij} - \mu)^T, \quad
    L_{\text{EigenCAM}} = v_1 A^k
    \label{eq:eq2}
\end{equation}
where $  A_{ij}  $ is the activation vector across channels at position $i,j$; $\mu$ is its mean over spatial locations, and $v_1$ is the first principal component (eigenvector of $C$). The map $L$ is the projection onto $v_1$.
\begin{figure}[ht!]
    \centering
    
    \subfloat[Real image]{%
        \includegraphics[width=0.42\columnwidth]{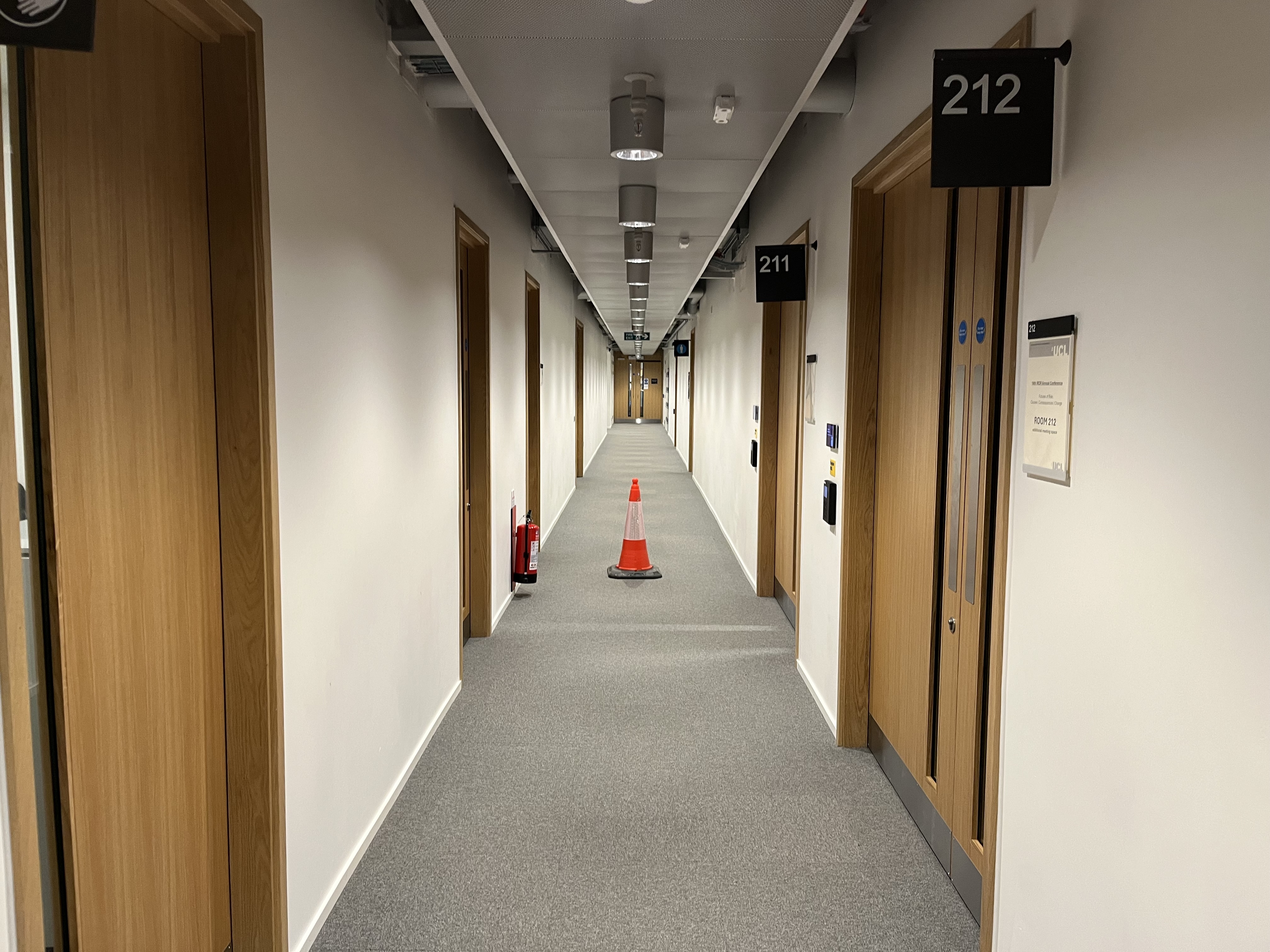}
        \label{subfig:image_real}
    } \quad
    \subfloat[Sim image]{%
        \includegraphics[width=0.42\columnwidth]{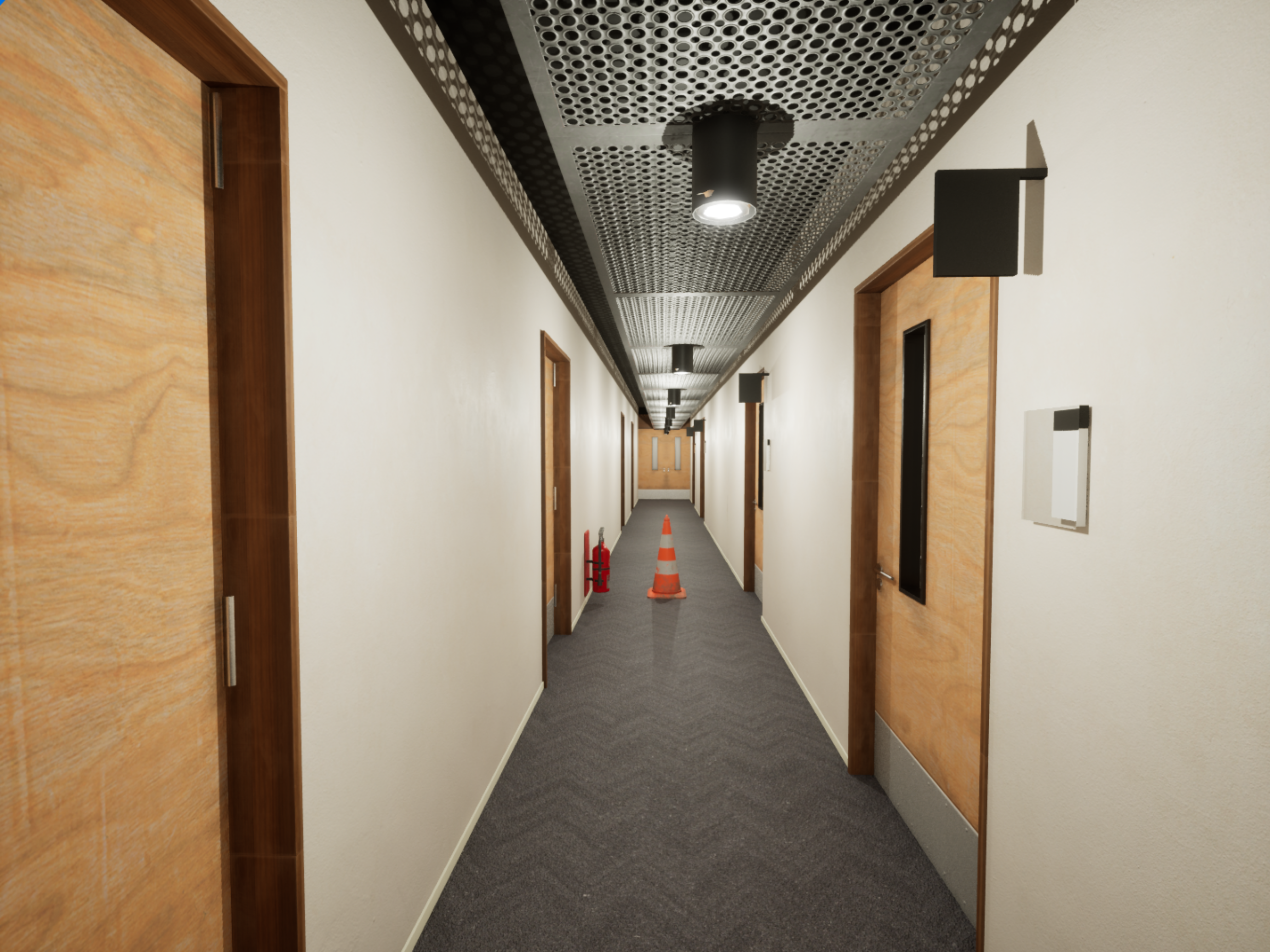}
        \label{subfig:image_sim}
    }\\
        \subfloat[GradCAM of CLIP on real image]{%
        \includegraphics[width=0.42\columnwidth]{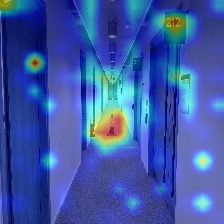}
        \label{subfig:gradcam_real}
    } \quad
    \subfloat[GradCAM of CLIP on Sim image]{%
        \includegraphics[width=0.42\columnwidth]{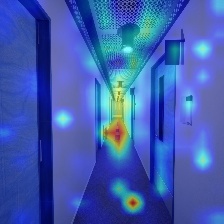}
        \label{subfig:gradcam_sim}
    } \\
    
    \subfloat[EigenCam of EfficientNet on real image]{%
        \includegraphics[width=0.42\columnwidth]{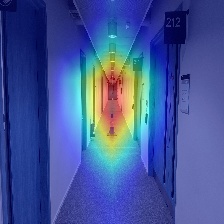}
        \label{subfig:eigencam_real}
    } \quad
    \subfloat[EigenCam of EfficientNet on Sim image]{%
        \includegraphics[width=0.42\columnwidth]{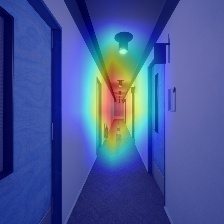}
        \label{subfig:eigencam_sim}
    }

    \caption{Comparison of real-world (left) and simulated (right) images, along with their corresponding Grad-CAM (CLIP) and EigenCAM (EfficientNet-B0) heatmaps. The highlighted class label is "cone".}
    \label{fig:eigencam_plot}%
    \vspace{-10.5pt}
\end{figure}

\textbf{Cosine Similarity} measures feature alignment between real and simulated embeddings:
\begin{equation}
    S_{\text{cosine}}(P, Q) = \frac{P \cdot Q}{\|P\| \|Q\|}
\end{equation}
where \( P \) and \( Q \) are feature vectors. Higher values indicate stronger similarity.

\textbf{KL Divergence} quantifies distribution differences:
\begin{equation}
    D_{\text{KL}}(P \| Q) = \sum_i P(i) \log \frac{P(i)}{Q(i)}
\end{equation}
Lower values indicate closer alignment. To compare models trained on real-world data, we use CLIP~\cite{radford2021learning} and the EfficientNet-B0~\cite{tan2019efficientnet} visual encoder used in ViNT~\cite{shah2023vint}. We first capture an image from the real world and recreate it within our simulator. Then both images are processed using Grad-CAM amd Eigen-Cam to generate attention heatmaps, as shown in Figure~\ref{fig:eigencam_plot}.
In both cases, the models primarily attend to the traffic cone in the center of the hallway as well as the end of the corridor, suggesting that the most salient features are preserved between the real and simulated environments. This alignment indicates that the models are focusing on the same high-level structures regardless of domain, reinforcing the realism of the simulated scene in terms of feature representation. The similarity in attention patterns also suggests that key objects remain distinguishable across domains, which is critical for transfer learning and sim-to-real applications. We further quantified the visual similarity between our hand-crafted simulations and real-world scenes by computing the KL divergence and cosine similarity for three scenes. These simulations were not built as photo-realistic replicas but as a proof of concept that our system can generate scenes visually representative of real conditions. As detailed in Table~\ref{tab:image_comparison_tab}, the Hallway scene, for example, yielded a KL divergence of 0.4679 and a cosine similarity of 0.7157, and an average of 0.4739 and 0.7094 in all scenes. These findings, along with our GradCAM and EigenCAM comparisons, confirm that our simulation is visually realistic enough for our evaluation purposes.

\begin{table}[ht!]
\centering
\resizebox{\columnwidth}{!}{
\begin{tabular}{ccccc}
\toprule
Metric  & Office Hallway              & Flat Hallway & Bedroom  & Avg \\ \midrule
KL divergence                        & 0.4679  &   0.4435                            & 0.5103         &   0.4739             \\
Cosine similarity                    &  0.7157                    & 0.7231                                      &     0.6895      &  0.7094   \\ \bottomrule
\end{tabular}
}
\caption{Image comparison metrics between real-world scenes and simulated recreations.}
\label{tab:image_comparison_tab}
\vspace{-10.5pt}
\end{table}


\subsection{Visual Navigation and SLAM Benchmarking}\label{Sec:Benchmark}
\begin{figure}[b!]
    \centering
    \subfloat[None]{%
        \includegraphics[width=0.315\columnwidth]{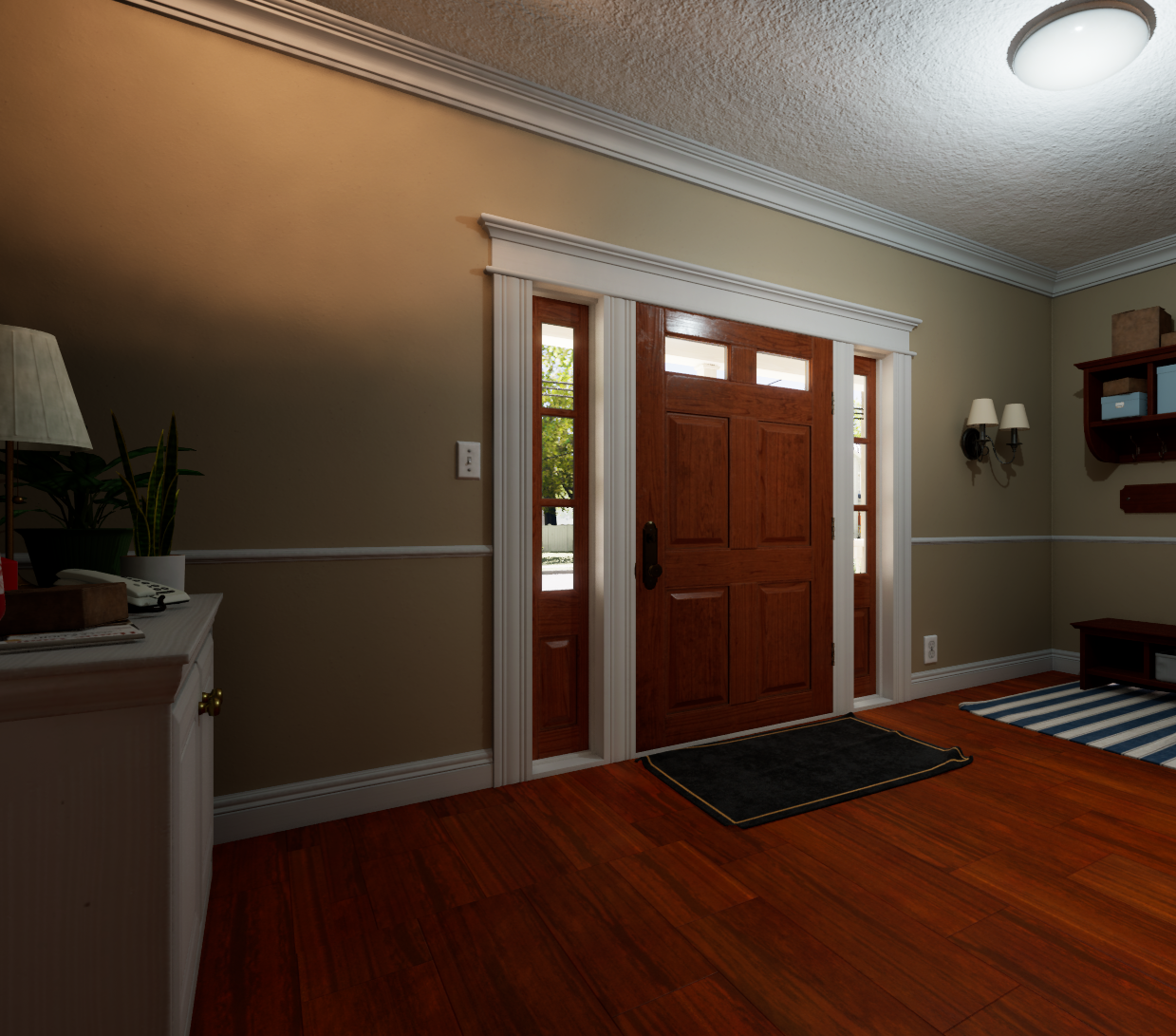}
        \label{subfig:adv_none}
    } 
    \subfloat[Minor]{%
        \includegraphics[width=0.315\columnwidth]{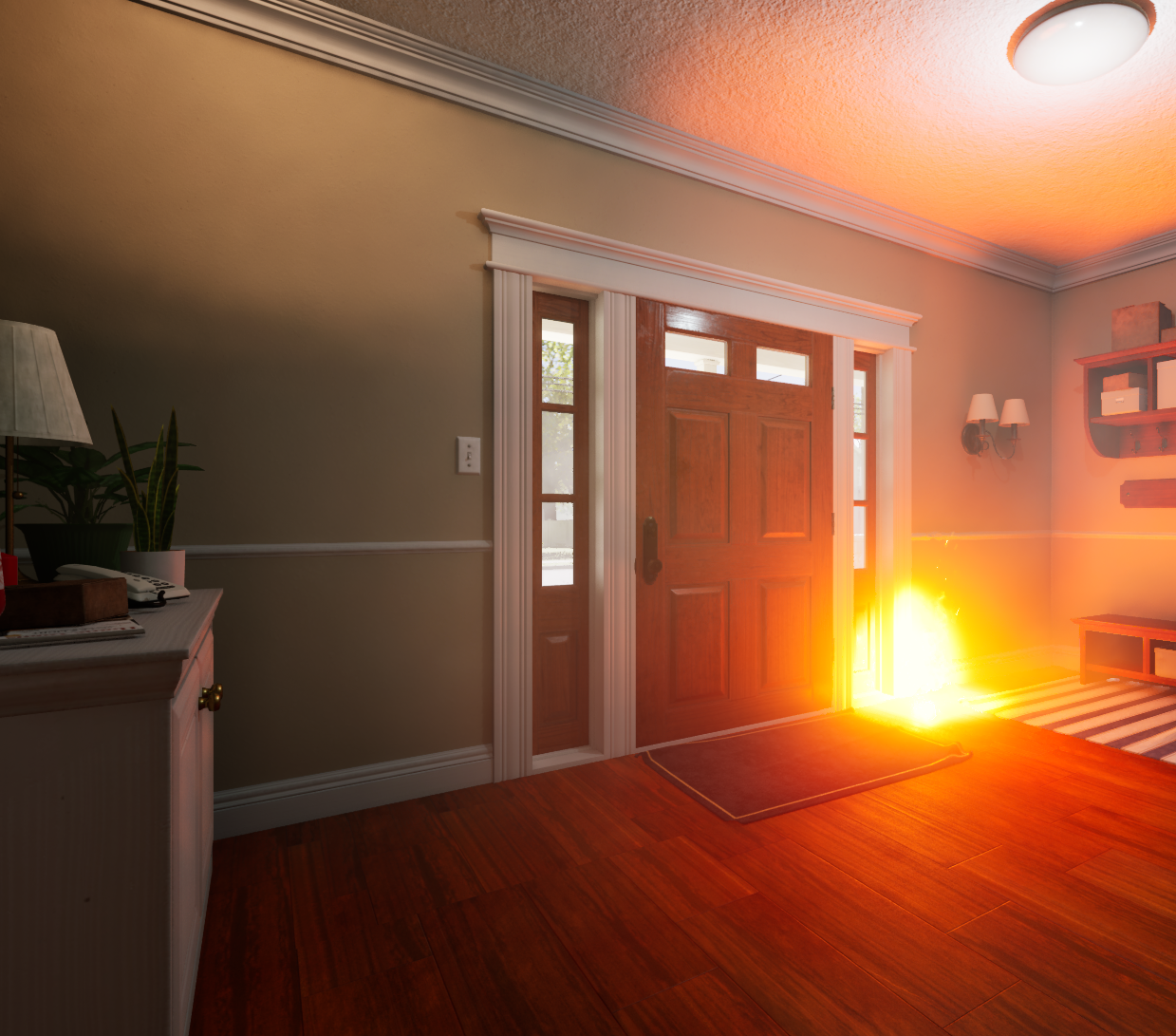}
        \label{subfig:adv_minor}
    }
     \subfloat[Major]{%
        \includegraphics[width=0.314\columnwidth]{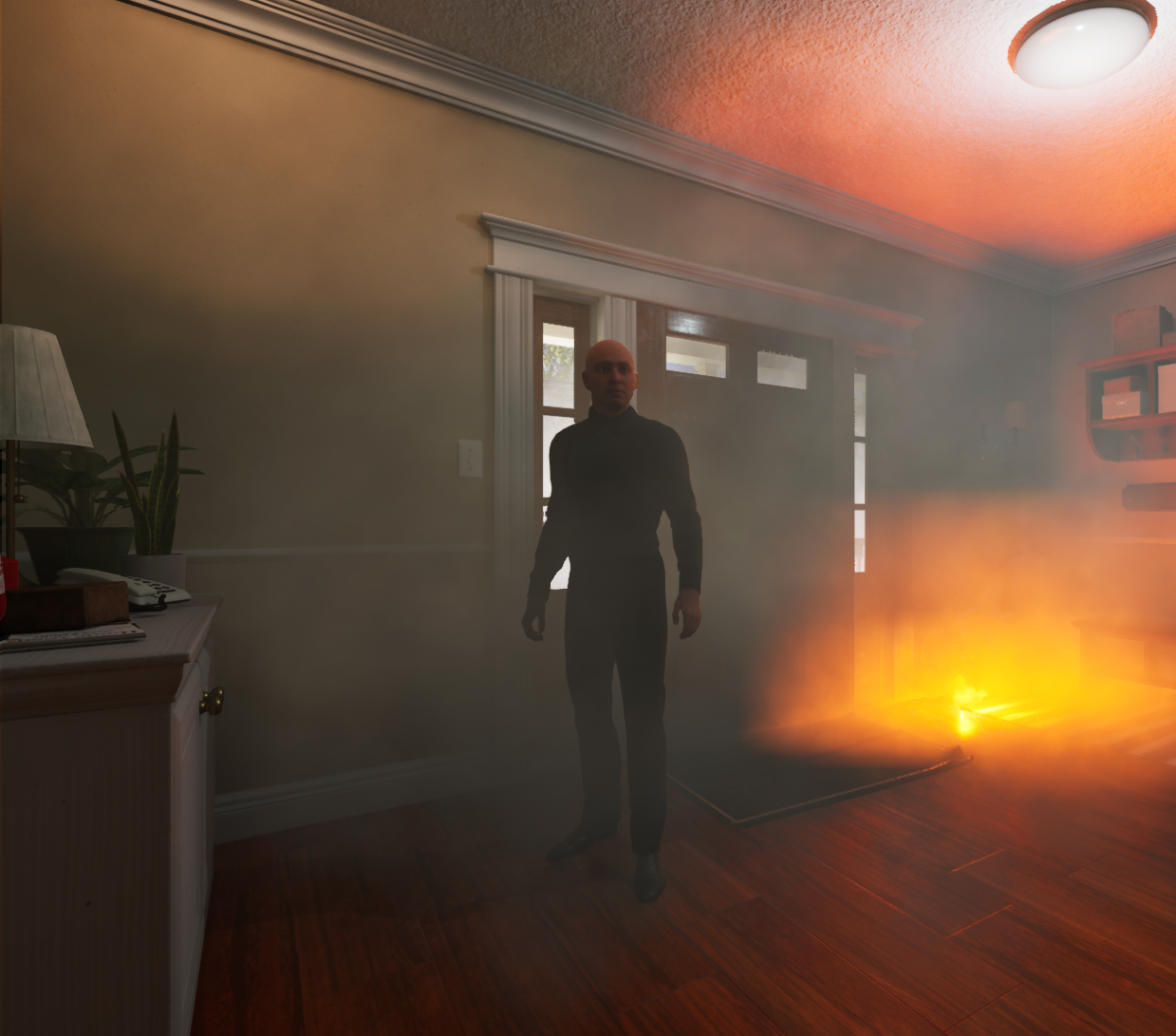}
        \label{subfig:adv_major}
    }
    \caption{Examples of the different adversity levels for the House Environment.}
    \label{fig:exp_adversity}%
    \vspace{-10.5pt}
\end{figure}
We benchmark standard methods to demonstrate simulator realism. We evaluated visual navigation (Experiment 1) with GNM~\cite{shah2023gnm} and ViNT~\cite{shah2023vint}, and visual SLAM (Experiment 2) with ORB-SLAM2~\cite{murORB2}, ORB-SLAM3~\cite{campos2021orb}, and MASt3R-SLAM~\cite{murai2024mast3r}. Each method was run 5 times per condition, averaging results across trials. Both experiments used the Unitree Go1 quadruped with its onboard camera and the Walk-These-Ways low-level controller for realistic deployment.
The benchmarks were conducted in two distinct environments—warehouse and residential house—under three adversity levels: none, minor and major, classified by human judgment. UE’s Niagara system simulated environmental effects (smoke, rain, pollution, snow), while dynamic lighting and NPC movements were added for major adversity. Figure~\ref{fig:example_photo_realistic_render} shows example environments and Figure~\ref{fig:exp_adversity} shows examples of adversity levels.

\subsubsection{Experiment 1: Visual Navigation}\label{subsubsec:vnav_benchmark}
For visual navigation benchmarking, the robot was initialized at identical starting locations across environments and adversity levels. Each method had a fixed duration to reach the goal using a pre-built image-based topological map. The warehouse and residential house were used for this experiment. Methods were not fine-tuned, relying on a simplified topological map approach from the authors' original repository~\cite{sridhar2024nomad}, which may limit performance~\cite{visualnav-transformer-issue5}.

\textbf{Success Weighted by Collision (SC).} Following~\cite{eftekhar2024one}, we adopt the SC metric:
\begin{equation}
    SC = \frac{1}{N} \sum_{i=1}^{N} S_i \frac{1}{1 + c_i},
\end{equation}
where \(S_i\) is the success indicator for episode \(i\), \(c_i\) is the number of collisions, and \(N\) is the total episodes. Higher SC values indicate fewer collisions in successful runs, while lower values reflect frequent or severe collisions. Alongside SC, we measure success rate, total collisions, time to goal, and goal distance.

\begin{table}[ht!]
    \scriptsize 
    \renewcommand{\arraystretch}{1.1} 
    \setlength{\tabcolsep}{3pt} 
\centering
\begin{tabular}{lllccccc}
\hline
Env & Adversity & Method & \begin{tabular}[c]{@{}c@{}}Success\\rate$\uparrow$\end{tabular} & Collisions$\downarrow$ & \begin{tabular}[c]{@{}c@{}}Time\\to Goal$\downarrow$\end{tabular} &  \begin{tabular}[c]{@{}c@{}}Goal\\Dist$\downarrow$\end{tabular} & \begin{tabular}[c]{@{}c@{}}Weighted\\SC$\uparrow$\end{tabular} \\ \hline
\multirow{6}{*}{House} & \multirow{2}{*}{Major}     & GNM    & 0.2          & 13         & 0                & 7.01      & 0.0154              \\
    &    & ViNT   & 0            & 9.2        & 0                & 8.41      & 0     \\\cline{2-8} 

    & \multirow{2}{*}{Minor}      & GNM    & 1            & 1.2        & 71.59            & 1.99      & 0.5333              \\        
    &      & ViNT   & 0.4          & 2.8        & 0                & 4.75      & 0.09  \\\cline{2-8} 
    
    & \multirow{2}{*}{None}     & GNM    & 1            & 1.8        & 147.86           & 1.99      & 0.3667              \\
    &     & ViNT   & 1            & 0.8        & 141.35           & 2.00      & 0.7333              \\
    \hline
\multirow{6}{*}{\begin{tabular}[c]{@{}c@{}}Ware-\\house\end{tabular}}     &  \multirow{2}{*}{Major}     & GNM    & 0            & 12.8       & 0                & 8.33      & 0                   \\
     &      & ViNT   & 0            & 6.4        & 0                & 7.96      & 0  \\\cline{2-8} 

     & \multirow{2}{*}{Minor}     & GNM    & 0.4          & 4          & 107.62           & 4.69      & 0.1067              \\
     &      & ViNT   & 0.2          & 6.6        & 54.91            & 6.61      & 0.0250 \\\cline{2-8} 

     & \multirow{2}{*}{None}     & GNM    & 0.8          & 3.4        & 102.86           & 2.53      & 0.1889              \\
     &    & ViNT   & 0.8          & 4.4        & 90.14            & 2.59      & 0.3667\\ \hline
\end{tabular}
\caption{Performance of different Visual navigation methods under various environments and varying adverse effects.}
\label{tab:visualnav_results}
\vspace{-10.5pt}
\end{table}
Table~\ref{tab:visualnav_results} presents the results, including Weighted SC to factor in collision penalties. State-of-the-art methods perform well under normal conditions, with ViNT generally outperforming GNM. Under minor adversity, GNM maintains a higher success rate but takes longer, while ViNT completes faster when successful. However, performance declines as environmental adversity increases (e.g., obstacles, smoke, rain), with both methods experiencing more collisions, longer times to goal, and lower SC. In severe adversity, neither method succeeds in the Warehouse, leading to zero SC. These failures are likely attributable to the dense, volumetric smoke simulated by the Niagara system, which caused persistent occlusion and aliasing of key visual features, preventing the vision-based planners from identifying a valid path. This suggests that these models struggle in extreme conditions due to limited training data, highlighting the need for training in diverse high-adversity conditions, which can be generated more effectively in simulation. The ability to create controlled but realistic adverse scenarios in a simulated environment provides a path to improve robustness in real-world deployments.

\subsubsection{Experiment 2: Visual SLAM}\label{subsubsec:vslam_Benchmark}
This experiment used our replay system, where a teleoperated robot followed predefined trajectories optimized for SLAM. Initial runs recorded robot poses and sensor data, which were later replayed with identical conditions but varying visual effects. Each SLAM method was then evaluated, and performance metrics were computed.

\textbf{Absolute Trajectory Error (ATE)} measures the global accuracy of an estimated trajectory by computing the RMSE between estimated positions $\hat{\mathbf{p}}_i$ and ground truth positions $\mathbf{p}_i$:

\begin{equation}
    \text{ATE} = \sqrt{\frac{1}{n} \sum_{i=1}^{n} \left(\mathbf{p}_i - \hat{\mathbf{p}}_i\right)^2}
\end{equation}
\textbf{Coverage} measures the ratio of the estimated trajectory length to the ground truth:
\begin{equation}
    \text{Coverage} = \frac{\sum_{i=1}^{n-1} \left\|\mathbf{p}_{i+1} - \mathbf{p}_i\right\|}{\sum_{i=1}^{n-1} \left\|\hat{\mathbf{p}}_{i+1} - \hat{\mathbf{p}}_i\right\|}
\end{equation}
Note that since monocular SLAM estimates trajectories with arbitrary scale, Umeyama's method~\cite{umeyama1991least} is applied for scale and pose correction before computing coverage. Poor tracking or alignment can result in coverage exceeding 100\%.

\textbf{Scaled ATE} adjusts ATE for localization failures:
\begin{equation}
    \text{Scaled ATE} = \frac{\text{ATE}}{\text{Coverage}}
\end{equation}

\begin{figure}[ht!]
    \centering
    \subfloat[Scaled ATE]{%
        \includegraphics[width=0.9 \columnwidth]{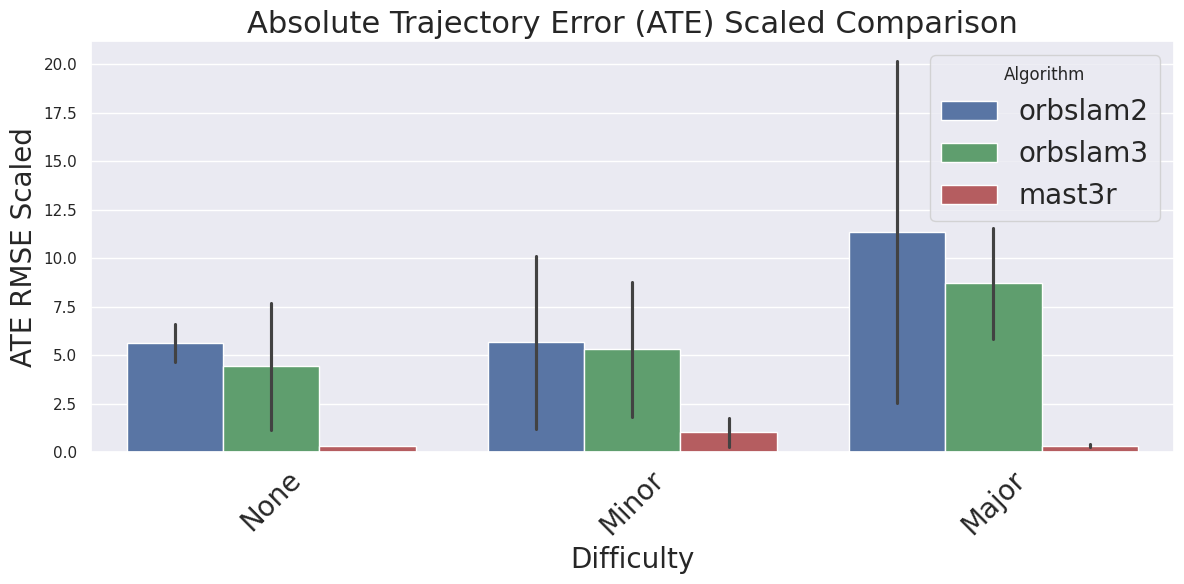}
        \label{subfig:APE_scaled_difficulty}
    } \\
    \subfloat[Coverage]{%
        \includegraphics[width=0.9 \columnwidth]{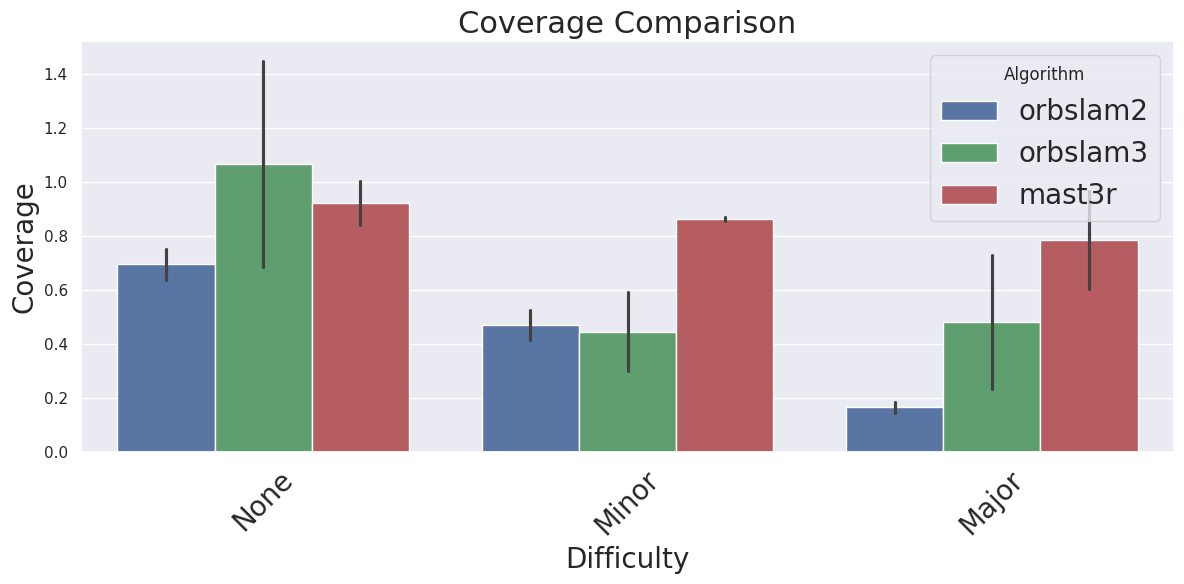}
        \label{subfig:Coverage_difficulty}
    }
    \caption{comparision of different Visual SLAM algorithms under different visual effects}
    \label{fig:VSLAM_bench}%
\end{figure}


\begin{table}[ht!]
    \centering
    \scriptsize 
    \renewcommand{\arraystretch}{1.1} 
    \setlength{\tabcolsep}{3pt} 
    \resizebox{\columnwidth}{!}{%
   \begin{tabular}{ll|ccc|ccc|ccc}
        \toprule
        \multirow{2}{*}{Env} & \multirow{2}{*}{Adver.} 
        & \multicolumn{3}{c|}{OrbSLAM2 Mono} & \multicolumn{3}{c|}{OrbSLAM3 Mono} 
        & \multicolumn{3}{c}{MASt3R-SLAM} \\
        \cmidrule(lr){3-5} \cmidrule(lr){6-8} \cmidrule(lr){9-11} 
        & & ATE$\downarrow$ & \begin{tabular}[c]{@{}c@{}}Scaled\\ATE$\downarrow$\end{tabular} & \begin{tabular}[c]{@{}c@{}}Cover-\\age$\uparrow$\end{tabular} 
        & ATE$\downarrow$ & \begin{tabular}[c]{@{}c@{}}Scaled\\ATE$\downarrow$\end{tabular} & \begin{tabular}[c]{@{}c@{}}Cover-\\age$\uparrow$\end{tabular}  
        & ATE$\downarrow$ & \begin{tabular}[c]{@{}c@{}}Scaled\\ATE$\downarrow$\end{tabular} & \begin{tabular}[c]{@{}c@{}}Cover-\\age$\uparrow$\end{tabular}  \\
        \midrule
        \multirow{3}{*}{\begin{tabular}[c]{@{}c@{}}Ware-\\house\end{tabular}} 
        & None  & 4.23  & 6.61  & 0.639  
                  & 5.29  & 7.72  & 0.685  
                  & 0.324  & 0.322  & 1.01 \\
         & Minor & 5.33  & 10.1  & 0.527  
                  & 5.21  & 8.80  & 0.592  
                  & 1.55   & 1.77  & 0.872  \\
         & Major & 2.96  & 20.2  & 0.147  
                  & 4.29  & 5.87  & 0.731  
                  & 0.252  & 0.417  & 0.605  \\
        \midrule
        \multirow{3}{*}{House} 
         & None  & 3.49  & 4.64  & 0.752  
                  & 1.69  & 1.17  & 1.45  
                  & 0.274  & 0.326  & 0.840  \\
         & Minor & 0.513 & 1.23  & 0.415  
                  & 0.547 & 1.81  & 0.302  
                  & 0.257  & 0.300  & 0.855  \\
         & Major & 0.475 & 2.53  & 0.188  
                  & 2.70  & 11.6  & 0.233  
                  & 0.275  & 0.284  & 0.966  \\
        \bottomrule
    \end{tabular}
    }
    \caption{Performance of Different VSLAM Methods under Varying Visual Effects.}
    \label{tab:vslam_results}
\end{table}

From Table~\ref{tab:vslam_results} and Fig.~\ref{fig:VSLAM_bench}, Visual SLAM performance degrades with increasing visual adversity, highlighting sensitivity to challenging conditions. ORB-SLAM2 struggles the most, with sharp declines in coverage and increased ATE, particularly in the warehouse environment. ORB-SLAM3 performs well in clean conditions but deteriorates significantly under adversity, suggesting reduced robustness to occlusions, lighting changes, and dynamic elements. MASt3R-SLAM maintains the lowest ATE and highest coverage across conditions, demonstrating superior resilience to visual disturbances.
Under severe visual adversity, all methods experience increased ATE and reduced coverage, highlighting the inherent challenges of maintaining reliable SLAM performance in dynamic or visually degraded environments. These replay-based evaluations provide a controlled, repeatable framework for assessing SLAM robustness across different levels of adversity, reinforcing the need for training and evaluation in diverse, high-adversity conditions to enhance real-world deployment reliability.

\subsubsection{Experiment 3: Real-world}
\begin{figure}[ht!]
    \centering
    \subfloat[Clean]{%
        \includegraphics[width=0.48\columnwidth]{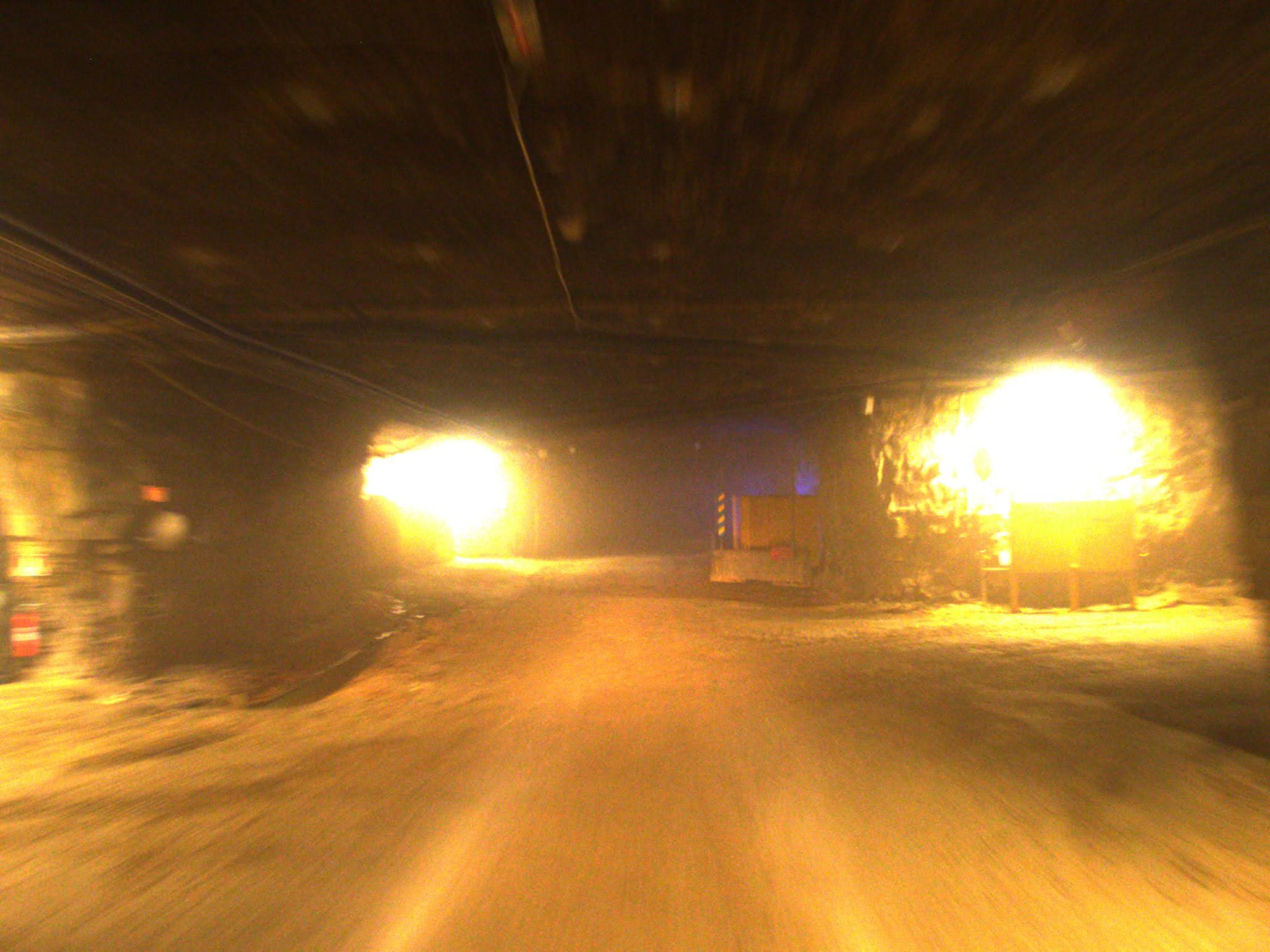}
        \label{subfig:realworld_clean}

    } 
    \subfloat[With smoke]{%
        \includegraphics[width=0.48\columnwidth]{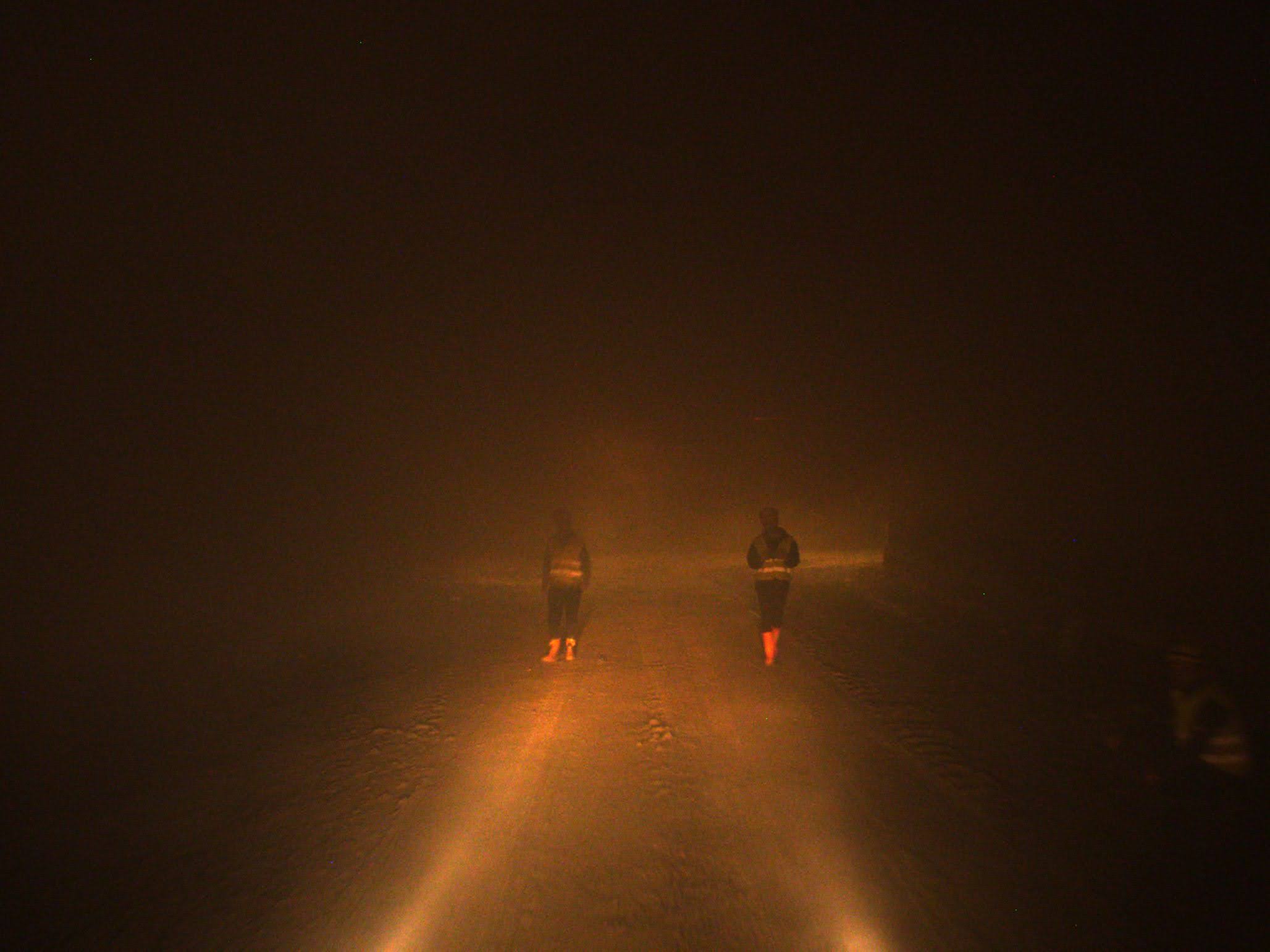}
        \label{subfig:realworld_smoky}
    }
    \caption{Examples of Realworld Tunnel data~\cite{kubelka2024we} used for evaluating VSLAM algorithm under adverse visual conditions.}
    \label{fig:exp_realword_adversity}%
\end{figure}

Visual SLAM algorithms were evaluated on realworld tunnel dataset~\cite{kubelka2024we} under two conditions: clear and smoky, where a section of the tunnel was obscured by smoke generated from a smoke machine. Examples of these conditions are shown in Fig.~\ref{fig:exp_realword_adversity}. The percentage of tracked frames under these conditions was measured for various visual SLAM algorithms, results can be seen in Tab.~\ref{tab:vslam_results_real}, where it can be seen that the tracking performance of all SLAM algorithm degrades significantly under smoky conditions, consistent with the findings from our simulated experiments.


\begin{table}[t!]
\centering
\begin{tabular}{c|c|ccc}
\toprule
\multirow{2}{*}{Env}  & \multirow{2}{*}{Adver.} & \multicolumn{3}{c}{Frames tracked $[\%]\uparrow$}                                                                                                                                                                        \\ \cmidrule(lr){3-5}
                      &                         & \multicolumn{1}{c|}{\begin{tabular}[c]{@{}c@{}}OrbSLAM2\\ Mono\end{tabular}} & \multicolumn{1}{c|}{\begin{tabular}[c]{@{}c@{}}OrbSLAM3\\ Mono\end{tabular}} & \begin{tabular}[c]{@{}c@{}}Mast3R\\ SLAM\end{tabular} \\         
\midrule

\multirow{2}{*}{Mine} & Clean                   & \multicolumn{1}{c|}{7.83}                                                    & \multicolumn{1}{c|}{59.6}                                                    & 100                                                   \\
                      & Smoky                   & \multicolumn{1}{c|}{0.00}                                                    & \multicolumn{1}{c|}{38.5}                                                    & 43.6 \\
\bottomrule           
\end{tabular}
\caption{Performance of Different VSLAM Methods under Varying Visual Effects in the Real-World.}
\label{tab:vslam_results_real}
\vspace{-10.5pt}
\end{table}

\subsubsection{Discussion}
Our experiments collectively demonstrate that current state-of-the-art visual navigation and SLAM algorithms, while proficient in ideal conditions, remain brittle to the types of realistic, adverse visual phenomena that are produced by our framework and that are encountered in the realworld. The observed performance degradation highlights a critical gap in current training and testing paradigms. Futhermore, because adverse real-world datasets are scarce, there is a reinforced need for high-fidelity simulators such as URL, which can generate controlled, repeatable, and diverse high-adversity scenarios to advance the development of more robust robotic perception systems.

%% file: 05_conclusion.tex

\section{CONCLUSION}\label{sec:conclusion}
This paper presents a high-fidelity robotics simulation framework that integrates advanced photorealistic rendering with precise physics modeling to support vision-based robotics research and benchmarking. Using Unreal Engine’s rendering capabilities and MuJoCo’s accurate physics, our system enables evaluation of robotic navigation and SLAM methods under diverse and adverse environmental conditions. Experimental results, such as utilizing the Replay System to test varied visual effects on identical trajectories, and Niagara-driven adversity benchmarks highlighting planner failures in volumetric smoke, confirm that our simulator provides a robust testbed for evaluating real-world robustness in controlled, repeatable scenarios.

Despite its strengths, the system has certain limitations, including the lack of deformable terrain simulation, one-way kinematic coupling for UE driven Actors, and the absence of validated sensor noise models for depth and LiDAR data. To address these limitations and further improve our simulator, we plan to focus on narrowing the sim-to-real perceptual gap by implementing procedural material randomization, generate large-scale synthetic datasets to enhance vision-based model training, introduce VR teleop for data generation, expand automated scene generation, and enable bilateral physical interactions between MuJoCo and UE. Overall, our platform serves as a robust foundation for benchmarking and data generation, bridging the gap between simulation and real-world robotics deployment.




